\documentclass{article}
\usepackage{spconf,amsmath,graphicx,hyperref}
\usepackage{xcolor}
\usepackage{subfigure}
\usepackage{makecell}
\usepackage{multirow}
\usepackage{booktabs}
\usepackage{kotex}

\title{BBPE16: UTF-16-based Byte-Level Byte-Pair Encoding for Improved Multilingual Speech Recognition}
\name{Hyunsik Kim\sthanks{Equal contribution.}, Haeri Kim\textsuperscript{*}, Munhak Lee, Kyungmin Lee}
\address{Samsung Research\\
\texttt{\{hyunsik777.kim, haeri.kim, mun-hak.lee, k.m.lee\}@samsung.com}}
\copyrightnotice{\parbox[t]{\textwidth}{\raggedright\tiny Copyright 2026 IEEE. Published in ICASSP 2026 - 2026 IEEE International Conference on Acoustics, Speech and Signal Processing (ICASSP), scheduled for 4-8 May 2026 in Barcelona, Spain. Personal use of this material is permitted. However, permission to reprint/republish this material for advertising or promotional purposes or for creating new collective works for resale or redistribution to servers or lists, or to reuse any copyrighted component of this work in other works, must be obtained from the IEEE. Contact: Manager, Copyrights and Permissions / IEEE Service Center / 445 Hoes Lane / P.O. Box 1331 / Piscataway, NJ 08855-1331, USA. Telephone: + Intl.\ 908-562-3966.}}
\begin{document}
\ninept
\maketitle
\thispagestyle{preprint}
\begin{abstract}

Multilingual automatic speech recognition (ASR) requires tokenization that efficiently covers many writing systems.
Byte‑level BPE (BBPE) using UTF‑8 is widely adopted for its language‑agnostic design and full Unicode coverage, but its variable‑length encoding inflates token sequences for non‑Latin scripts, such as Chinese, Japanese, and Korean (CJK).
Longer sequences increase computational load and memory use.
We propose BBPE16, a UTF‑16-based BBPE tokenizer that represents most modern scripts with a uniform 2‑byte code unit.
BBPE16 preserves BBPE's language‑agnostic properties while substantially improving cross-lingual token sharing.
Across monolingual, bilingual, and trilingual ASR, and in a multilingual continual-learning setup, BBPE16 attains comparable or better accuracy; for Chinese, it reduces token counts by up to 10.4\% and lowers decoding iterations by up to 10.3\%.
These reductions speed up fine‑tuning and inference and decrease memory usage, making BBPE16 a practical tokenization choice for multilingual ASR.
\end{abstract}
\section{Introduction}
\label{sec:intro}

As automatic speech recognition (ASR) has advanced, attention has shifted from single-language systems to multilingual ASR.
The goal is not only higher accuracy~\cite{8461972} but also broader language coverage—ranging from dozens~\cite{pratap20c_interspeech} to, in some cases, hundreds of languages~\cite{zhang2023google, 10.5555/3722577.3722674}.
To achieve this breadth, the model must have an output vocabulary capable of representing a wide variety of writing systems; therefore, tokenization becomes a primary design consideration.

While character- or subword-level byte-pair encoding (BPE)~\cite{sennrich-etal-2016-neural} is often sufficient for monolingual ASR, multilingual setups commonly adopt byte-level BPE (BBPE) to ensure full Unicode coverage and to avoid language-specific preprocessing.
Contemporary BBPE implementations are predominantly UTF-8–based~\cite{8682674}, which brings practical advantages such as ASCII compatibility and robustness.
However, UTF-8 also introduces friction in cross-script settings.
Its variable-length encoding yields uneven token boundaries across languages and, for non-Latin scripts such as Chinese, Japanese, and Korean (CJK), inflates sequence length.
This lengthening increases computation and memory use, which in turn raises model and runtime complexity.
These effects make efficient, scalable multilingual ASR more challenging, which motivates alternative BBPE designs that better align with cross-lingual settings.

In real-world deployments, multilingual ASR systems are initially pretrained on massive multilingual corpora and then undergo continual learning---i.e., incremental adaptation to new domains, dialects, or underrepresented languages while preserving prior knowledge---to maintain or improve performance in specific target settings.
In such settings, models must incorporate new data without catastrophic forgetting of previously learned languages.
Against this backdrop, continual learning for multilingual ASR has become an active area of research~\cite{10.1109/TASLP.2024.3487410, kwok24_interspeech}.
These continual-learning scenarios impose strong requirements on tokenizer design: the tokenizer should facilitate efficient cross-lingual sharing and maintain stability as new languages or domains are incrementally added.

In this work, we propose BBPE16, a UTF-16-based BBPE tokenizer that addresses limitations of UTF-8-based BBPE while maintaining compatibility with existing infrastructure.
UTF-16 provides a uniform 2-byte representation for the vast majority of characters in the Basic Multilingual Plane (BMP), which includes most modern scripts, including CJK characters.
Importantly, in our trilingual setup BBPE16 exhibits markedly superior cross‑lingual token sharing capability, generating tokens shared among English, Korean, and Chinese, whereas BBPE yields none.
Our key contributions are:
\begin{itemize}
    \item A novel UTF-16-based BBPE tokenizer that achieves comparable ASR performance with significantly improved token efficiency and cross-lingual token sharing
    \item Up to 10.4\% token reduction for Chinese in a multilingual continual-learning scenario, alongside reductions in decoding iterations, yielding practical training and inference speedups
\end{itemize}

\section{Background}
\label{sec:bckg}
\subsection{Byte-Level BPE (BBPE)}
Modern Transformer~\cite{NIPS2017_3f5ee243}-based models (e.g., BERT~\cite{devlin-etal-2019-bert}, GPT-3~\cite{NEURIPS2020_1457c0d6}) are trained on hundreds of billions of tokens and contain tens of billions of parameters.
At this scale, word-level vocabularies suffer from severe sparsity and vocabulary-size explosion problems.
Subword methods such as BPE~\cite{sennrich-etal-2016-neural} and SentencePiece~\cite{kudo-richardson-2018-sentencepiece} reduce sparsity while still capturing a wide range of morphological variations.
However, classic BPE still requires language-specific preprocessing, such as word tokenizers and whitespace handling.
When processing multilingual corpora or corpora that contain a substantial number of low‑frequency characters, maintaining language‑specific preprocessing rules is cumbersome.

BBPE first converts the input text into a byte sequence and then applies BPE on that sequence.
This makes the tokenizer language-agnostic: as long as the text is encoded in UTF-8, any Unicode character can be processed by the same pipeline. 
BBPE has several advantages over conventional BPE: in multilingual settings, it distributes token usage more uniformly across the corpus and yields a larger set of tokens shared among languages~\cite{Wang_Cho_Gu_2020}.
Moreover, because BBPE's vocabulary encompasses all possible byte values, it inherently eliminates out‑of‑vocabulary tokens.

\subsection{Limitations of UTF-8-based BBPE}
While UTF-8 has become the de facto standard for text encoding due to its backward compatibility with ASCII and error resilience, it introduces several challenges for BBPE.

\noindent
\textbf{Variable-length encoding}:
Characters require 1-4 bytes depending on their Unicode code point, leading to inconsistent tokenization patterns.
To mitigate the discrepancy that arises when constructing BBPE for English–Chinese bilingual data caused by the variable‑length nature of the encoding, \cite{9747842} introduced a length penalty and an alphabet penalty.

\noindent
\textbf{Susceptibility to decoding errors}:
While UTF-8 enables strict validation by rejecting invalid byte sequences, this also makes it more susceptible to decoding errors.
To address these decoding errors, several studies have applied a dynamic‑programming algorithm~\cite{8682674, Wang_Cho_Gu_2020}, while another line of work explored a solution based on vector quantization~\cite{10832272}.

Although prior research has explored different approaches to mitigate the limitations of UTF-8-based BBPE, this work introduces a novel tokenizer that addresses these shortcomings through a simple modification.

\section{BBPE16: UTF-16-based Tokenization}
\label{sec:method}

\subsection{Motivation for a UTF-16-based BBPE}
\label{subsec:motivation}
UTF-16 uses 16-bit (2-byte) code units as its fundamental unit.
Characters in the BMP, which spans U+0000-U+FFFF, are represented with a single code unit (2 bytes).
The BMP includes most modern scripts, such as (1) Latin characters, (2) CJK unified ideographs, (3) Hangul syllables, and (4) major scripts like Arabic, Hebrew, and Devanagari.
This results in a language-agnostic and uniform 2-byte representation for each character, whereas UTF-8 employs a variable-length byte encoding per character.

\vspace{-0.3cm}
\begin{table}[htb]
\centering
\caption{Comparison of UTF-8 and UTF-16 (little-endian) encodings for the Korean Hangul syllable `한'}
\label{tab:utf-8}
\begin{tabular}{|c|c|c|c|}
\hline
\multicolumn{4}{|c|}{\textbf{Example: 한 (U+D55C)}} \\
\hline
\textbf{Encoding} & \textbf{Byte 1} & \textbf{Byte 2} & \textbf{Byte 3} \\
\hline
\multirow{2}{*}{\textbf{UTF-8}} & \underline{\textbf{\texttt{1110}}}\texttt{1101} & \underline{\textbf{\texttt{10}}}\texttt{010101} & \underline{\textbf{\texttt{10}}}\texttt{011100} \\
& (\texttt{ED}) & (\texttt{95}) & (\texttt{9C}) \\
\hline
\multirow{2}{*}{\textbf{UTF-16}} & \texttt{01011100} & \texttt{11010101} & \multirow{2}{*}{-} \\
& (\texttt{5C}) & (\texttt{D5}) & \\
\hline
\end{tabular}
\end{table}
\vspace{-0.2cm}

Table~\ref{tab:utf-8} compares UTF-8 and UTF-16 encoding for non-Latin scripts, focusing on CJK characters, with the Korean Hangul syllable `한' shown as a representative example.
In this study, we assume the use of UTF-16 encoding in little-endian format and thus we ignore the Byte Order Mark (BOM). 
Bold and underlined bits (e.g., \underline{\textbf{\texttt{1110}}}) indicate prefix bits prescribed by the UTF-8 encoding specification, while regular bits indicate data bits.
In UTF-8, multiple bits are used as a length-indicating prefix, while in UTF-16 all bits encode character data.
Because UTF-16 eliminates the overhead of length-prefix bits, a UTF-16-based tokenization scheme in turn allows the vocabulary to be constructed more compactly and efficiently than UTF-8-based BBPE.

\subsection{BBPE16: UTF-16-based BBPE}
We propose BBPE16, a BBPE tokenizer that operates on UTF-16 rather than the conventional UTF-8 encoding. BBPE16 follows the standard BPE merge algorithm, but all merges are performed on the byte sequences derived from UTF-16 code units. The processing pipeline of BBPE16 is outlined below:

\begin{enumerate}
    \item \textbf{Encoding:} Convert the input text from UTF-8 to UTF-16 (little-endian)
    \item \textbf{Byte extraction:} Obtain the raw UTF-16 bytes and discard the BOM
    \item \textbf{BPE training:} Learn merge rules over the UTF-16 byte sequences
    \item \textbf{Tokenization:} Apply the learned merges to new UTF-16 encoded texts
    \item \textbf{Decoding:} Reconstruct the UTF-16 text from tokens and convert back to UTF-8 text
\end{enumerate}

As described in Section \ref{subsec:motivation}, we adopt UTF-16 little-endian encoding, and therefore the BOM is discarded as part of the preprocessing step. 
Input and output remain UTF-8 encoded, ensuring compatibility with existing systems. Only the internal tokenization operates on UTF-16, making BBPE16 a drop-in replacement for UTF-8-based BBPE.

\section{Experimental Settings}
\label{sec:experiments}

\subsection{Datasets}
We evaluated BBPE16 across multiple language settings using LibriSpeech~\cite{librispeech} for English, KsponSpeech~\cite{KsponSpeech} for Korean, and AISHELL-1~\cite{8384449} for Chinese.
For the continual-learning experiments we used Wall Street Journal (WSJ)~\cite{paul-baker-1992-design} for English, Zeroth-Korean~\cite{zeroth2019} for Korean, and Common Voice~\cite{ardila-etal-2020-common} Chinese dataset for Chinese.

For LibriSpeech corpus, we used speed perturbation (factors 0.9×, 1.0×, and 1.1×), generating 843,723 total training utterances from original 281,241.
Speed perturbation was applied only to English monolingual training, not to bilingual or trilingual settings.
Text was normalized using the ESPnet~\cite{watanabe18_interspeech} pipeline with only essential tokens and apostrophes retained.
For KsponSpeech, pronunciation notation was used with English letters converted to Korean phonetics.
For Zeroth-Korean, we randomly split the training data into 90\% for training and 10\% for development.
For Common Voice Chinese, embedded English letters were uppercased.
All datasets were filtered to exclude utterances over 30 seconds or with empty normalized text.
This removed 496 utterances from KsponSpeech and 9 from WSJ.

\subsection{Model Architecture and Inference Settings}
We employed ESPnet~\cite{watanabe18_interspeech} to train attention-based encoder-decoder (AED) models with E-Branchformer~\cite{10022656} encoders.
The encoder consists of 17 blocks with 512-dimensional outputs, 1,024 linear units, and 8 attention heads with relative positional encoding.
The decoder uses 6 Transformer~\cite{NIPS2017_3f5ee243} blocks with 2,048 linear units and 8 attention heads.
All models used identical architectures, differing only in the tokenization strategy.
All models---except the continual-learning model described in Section \ref{subsec:adaptation}---were trained for 80 epochs.
The continual-learning model, which starts from the trilingual model, was trained for 30 epochs.
Decoding used beam search with beam size 4.

\subsection{Tokenizer Configurations}
We compared three tokenization approaches:
\begin{itemize}
    \item \textbf{BPE:} Standard character-level BPE
    \item \textbf{BBPE:} Byte-level BPE on UTF-8 encoding
    \item \textbf{BBPE16:} Our proposed UTF-16-based BBPE
\end{itemize}

\section{Results and Analysis}
\label{sec:results}

\subsection{Monolingual and Bilingual Scenarios}
\subsubsection{ASR Performance}
To verify that BBPE16 introduces no degradation in ASR performance, we trained and evaluated monolingual English, monolingual Korean, and bilingual (English \& Korean) ASR models using a variety of tokenizers, and compared the resulting recognition accuracies.

\vspace{-0.3cm}
\begin{table}[htb]
\centering
\caption{WER (\%) comparison across tokenizers in monolingual and bilingual scenarios}
\label{tab:wer_results}
\begin{tabular}{ccc|ccc}
\toprule
\textbf{Language} & \multicolumn{1}{c}{\begin{tabular}[c]{@{}c@{}}\textbf{Vocab}\\\textbf{Size}\end{tabular}} & \textbf{Test Set} & \textbf{BPE} & \textbf{BBPE} & \textbf{BBPE16} \\
\midrule
\multirow{2}{*}{English}   & \multirow{2}{*}{1000} & test-clean & 2.1    & 2.2     & 2.1       \\
                           &                       & test-other & 4.8    & 4.7     & 4.6       \\
\midrule
\multirow{2}{*}{Korean}    & \multirow{2}{*}{3000} & Eval-clean & 18.5    & 18.7     & 18.6       \\
                           &                       & Eval-other & 21.5    & 21.8     & 22.0      \\
\midrule
\multirow{4}{*}{\parbox[c]{1.4cm}{\centering Bilingual\\(En \& Ko)}} & \multirow{4}{*}{5000} & test-clean & 2.5    & 2.7     & 2.6       \\
                           &                       & test-other & 5.8    & 6.1     & 6.0       \\
                           &                       & Eval-clean & 19.0    & 18.9     & 19.1       \\
                           &                       & Eval-other & 22.1    & 22.6     & 22.2       \\
\bottomrule
\end{tabular}
\end{table}
\vspace{-0.2cm}

Table~\ref{tab:wer_results} presents word error rate (WER) comparisons across different settings.
In both monolingual and bilingual settings, BBPE16 shows no significant performance difference compared with BPE and BBPE on any test set.

\subsection{Trilingual Scenario}
To understand the cross-lingual capabilities and the tokenizer efficiency of BBPE16, we compared trilingual tokenizers trained on combined datasets (LibriSpeech, KsponSpeech, and AISHELL-1).
We used a vocabulary size of 7,000 for trilingual tokenizers.
\subsubsection{Shared Token Analysis}
\label{subsec:common_token}
We analyzed token sharing patterns across English, Korean and Chinese with trilingual tokenizers.
Table~\ref{tab:common_tokens} shows the number of common tokens shared between language pairs for BBPE and BBPE16.

\vspace{-0.3cm}
\begin{table}[htb]
\centering
\caption{Number of shared tokens across language pairs in trilingual tokenizers}
\label{tab:common_tokens}
\begin{tabular}{lcc}
\toprule
\textbf{Language Pair} & \textbf{BBPE} & \textbf{BBPE16} \\
\midrule
English-Korean & 0 & 42 \\
Korean-Chinese & 95 & 573 \\
Chinese-English & 0 & 55 \\
All 3 Languages & 0 & 42 \\
\bottomrule
\end{tabular}
\end{table}
\vspace{-0.2cm}

The results demonstrate BBPE16's remarkable advantage in creating cross-lingual token compatibility.
Notably, BBPE yields no shared tokens for the English-Korean pair, the Chinese-English pair, and for the three-language set combined, whereas BBPE16 produces at least 42 shared tokens in each of those three cases\footnote{The byte sequence "ĠĀW" and "ĠĀL" are examples of cross-lingual shared tokens, appearing in English, Korean, and Chinese.}.
BBPE16 shows substantial improvements in Korean-Chinese token sharing (95→573) and notable gains in other language pairs, demonstrating its effectiveness for multilingual tokenization.

\subsubsection{Token Count Analysis}
Using the same trilingual tokenizer, we evaluated compression efficiency across individual languages.
Table~\ref{tab:trilingual_tokens} presents detailed token statistics.

\vspace{-0.3cm}
\begin{table}[htb]
\centering
\caption{Average token counts per utterance for each language using trilingual tokenizers, with the last column showing the relative reduction of BBPE16 compared to BBPE}
\label{tab:trilingual_tokens}
\begin{tabular}{lcccc}
\toprule
\textbf{Language} & \textbf{BPE} & \textbf{BBPE} & \textbf{BBPE16} & \textbf{BBPE16 vs BBPE} \\
\midrule
English & 76.5 & 45.4 & 45.2 & -0.4\% \\
Korean & 23.5 & 16.5 & 16.3 & -1.2\% \\
Chinese & 22.3 & 19.5 & 18.6 & -4.6\% \\
\bottomrule
\end{tabular}
\end{table}
\vspace{-0.2cm}

This analysis reveals the advantages of BBPE16, showing slight improvement for English (-0.4\%), modest improvement for Korean (-1.2\%), and significant gains for Chinese (-4.6\%) compared to BBPE.
Notably, BPE allocates most of its limited merged‑token budget to cover the entire Korean and Chinese character ranges, which leaves relatively few tokens available for English.
As a result, BPE produces a large number of tokens for English text while still remaining competitive for Korean and Chinese.
The efficiency of BBPE16 directly leads to faster training and lower memory usage.

\subsubsection{Vocabulary Coverage Analysis}
Beyond token efficiency, vocabulary coverage (the proportion of the total vocabulary utilized by a given language) is a critical metric for tokenizer effectiveness.
Higher vocabulary coverage reduces wasted token slots, enabling more efficient token space usage and better utilization of the model’s embedding parameters in multilingual settings.
Table~\ref{tab:vocab_usage} illustrates the percentage of vocabulary utilized by each trilingual tokenizer across languages.

\vspace{-0.3cm}
\begin{table}[htb]
\centering
\caption{Vocabulary coverage (\%) across trilingual tokenizers and languages}
\label{tab:vocab_usage}
\begin{tabular}{lccc}
\toprule
\textbf{Language} & \textbf{BPE} & \textbf{BBPE} & \textbf{BBPE16} \\
\midrule
English & 4.5 & 44.1 & 48.1 \\
Korean & 35.0 & 39.9 & 43.0 \\
Chinese & 60.5 & 14.7 & 17.8 \\
\bottomrule
\end{tabular}
\end{table}
\vspace{-0.2cm}

BBPE16 consistently achieves higher vocabulary coverage across all languages, with improvements of 4.0\% for English, 3.1\% for Korean, and 3.1\% for Chinese compared to BBPE.
The superior vocabulary coverage of BBPE16 indicates that its token inventory is used more efficiently across all three languages.

In the trilingual experiments, the English vocabulary usage proportion under BPE is markedly lower than that of Korean and Chinese. This is because English contains far fewer distinct characters, whereas the token budget must accommodate the far larger and more diverse character sets of Korean and Chinese.

As discussed in Section~\ref{subsec:common_token}, BBPE16’s cross-lingual sharing leads the aggregated per-language coverage to exceed 100\%.
In contrast, many pre-allocated byte tokens in BBPE are never selected during training, so its aggregated coverage remains below 100\%.

\subsection{Continual-Learning Scenario}
\label{subsec:adaptation}
We evaluated the effectiveness of BBPE16 not only in the trilingual setting but also in a continual-learning scenario using trilingual tokenizers.
These experiments demonstrate that BBPE16 offers clear advantages when incorporating new data.

We further evaluated BBPE16 in a continual-learning setup by fine-tuning the trilingual model on the base corpus (LibriSpeech, KsponSpeech, and AISHELL-1) plus three additional datasets: WSJ (English), Zeroth-Korean (Korean), and Common Voice Chinese (Chinese).

For this scenario, we excluded the standard BPE tokenizer from the experiments because its out‑of‑vocabulary handling makes it unsuitable for this particular setting.

\noindent
\textit{Note:} From this point, Zeroth‑Korean is referred to as Zeroth, and Common Voice Chinese as CVC.

\subsubsection{ASR Performance}

\begin{table}[t]
\centering
\caption{Performance comparison between BBPE and BBPE16 in trilingual and continual-learning scenarios, with WER (\%) for English and Korean and CER (\%) for Chinese}
\label{tab:adapt_wer}
\begin{tabular}{lcccc}
\toprule
\textbf{Dataset} & \multicolumn{2}{c}{\textbf{Trilingual}} & \multicolumn{2}{c}{\textbf{Continual-Learning}} \\
\cmidrule(lr){2-3} \cmidrule(lr){4-5}
 & \textbf{BBPE} & \textbf{BBPE16} & \textbf{BBPE} & \textbf{BBPE16} \\
\midrule
\multicolumn{5}{l}{\textit{Base datasets}} \\
\midrule
LibriSpeech & & & & \\
\quad test-clean & 2.7 & 2.6 & 2.6 & 2.5 \\
\quad test-other & 6.1 & 6.1 & 5.8 & 5.8 \\
KsponSpeech & & & & \\
\quad Eval-clean & 18.7 & 19.0 & 18.7 & 18.7 \\
\quad Eval-other & 22.2 & 22.4 & 22.0 & 21.9 \\
AISHELL-1 & 5.9 & 5.7 & 5.6 & 5.6 \\
\midrule
\multicolumn{5}{l}{\textit{Additional datasets}} \\
\midrule
WSJ & 10.7 & 10.8 & 4.8 & 4.2 \\
Zeroth & 76.0 & 47.7 & 7.6 & 7.5 \\
CVC & 245.7 & 273.9 & 15.6 & 15.6 \\
\bottomrule
\end{tabular}
\end{table}

The ASR performance comparison in the continual-learning scenario is presented in Table~\ref{tab:adapt_wer}.
The values in Table~\ref{tab:adapt_wer} are WER for English and Korean datasets, and character error rate (CER) for Chinese dataset.
Continual learning substantially improves performance on all additional datasets over the trilingual baseline.
Overall, BBPE16 delivers performance comparable to that of BBPE—matching BBPE on CVC while offering modest improvements on WSJ and Zeroth.

\subsubsection{Token Count Analysis}
\label{subsubsec:token_count}

Table~\ref{tab:adaptation} shows the token count comparison for each additional dataset in the continual-learning scenario.
BBPE16 demonstrates consistent advantages across all languages, with no change for English but improvement for the Korean dataset, and substantial advantages for the Chinese dataset, achieving a 10.4\% token reduction on the CVC compared to BBPE.
This significant reduction in the Chinese dataset leads to computational savings during continual learning for Chinese.

\begin{table}[t]
\centering
\caption{Average token counts per utterance for each additional dataset in the continual-learning scenario, with the last column showing the relative reduction of BBPE16 compared to BBPE}
\label{tab:adaptation}
\begin{tabular}{lccc}
\toprule
\textbf{Dataset} & \textbf{BBPE} & \textbf{BBPE16} & \textbf{BBPE16 vs BBPE} \\
\midrule
WSJ & 28.7 & 28.7 & 0.0\% \\
Zeroth & 37.0 & 36.7 & -0.8\% \\
CVC & 28.9 & 25.9 & -10.4\% \\
\bottomrule
\end{tabular}
\end{table}

\subsubsection{Inference Efficiency}

In Table~\ref{tab:adapt_token}, we report the average number of decoding iterations per utterance, defined as the number of decoding steps required to emit the end-of-sequence token under beam search.
Consistent with the training token counts, BBPE16 achieves reductions in the decoding iterations across all languages: modest improvements for English (-0.4\%) and Korean (-0.8\%), and substantial efficiency gains for Chinese (-10.3\%).
This demonstrates that BBPE16 offers consistent efficiency improvements across all languages, with particularly significant gains for Chinese both during training and inference.

\begin{table}[t]
\centering
\caption{Average decoding iterations per utterance for each test set in the continual-learning scenario, with the last column showing the relative reduction of BBPE16 compared to BBPE}
\label{tab:adapt_token}
\begin{tabular}{lccc}
\toprule
\textbf{Test Set} & \textbf{BBPE} & \textbf{BBPE16} & \textbf{BBPE16 vs BBPE} \\
\midrule
WSJ    & 27.3 & 27.2 & -0.4\%\\
Zeroth & 36.8 & 36.5 & -0.8\%\\
CVC    & 27.3 & 24.5 & -10.3\%\\
\bottomrule
\end{tabular}
\end{table}

\section{Discussion}
\label{sec:discussion}

In many practical scenarios, the training corpus is dominated by large-scale English data, while languages such as CJK are represented with relatively smaller amounts of data.
Subsequently, it is common to augment the training set with additional data to improve performance in non-English languages.
In such cases, BBPE16 offers advantages over conventional BBPE by achieving more efficient tokenization through cross-lingual token sharing and 2-byte encoding.
Furthermore, BBPE16 retains compatibility with existing BBPE since it differs only in replacing UTF-8 with UTF-16 for text encoding.
Beyond the scope of speech recognition, this property also makes BBPE16 applicable to broader language modeling tasks, including large language models.

\section{Conclusion}
\label{sec:conclusion}

We present BBPE16, a UTF‑16-based BBPE tokenizer that overcomes the key limitations of BBPE in multilingual settings.
By exploiting UTF‑16’s uniform code‑unit representation, BBPE16 reduces token counts by up to 10.4\% in a multilingual continual-learning scenario, alleviating the efficiency bottlenecks of multilingual ASR.

Rather than sacrificing English token efficiency, BBPE16 improves tokenization across all languages, yielding better cross‑lingual token sharing and lower computational cost.

As multilingual ASR becomes increasingly essential, BBPE16 offers a practical, high‑efficiency solution that especially benefits non-Latin scripts like CJK while maintaining competitive or superior performance across diverse multilingual scenarios.

\vfill\pagebreak

\bibliographystyle{IEEEtran}
\bibliography{refs}

\end{document}